\documentclass[a4paper]{article}

\usepackage{INTERSPEECH2021}
\usepackage{microtype}
\usepackage{float}
\usepackage{url}
\usepackage{balance}
\usepackage{graphics}
\usepackage{epstopdf}
\usepackage{subfig}
\usepackage{hyperref}
\usepackage{xstring}
\usepackage{xcolor}
\usepackage{tabularx}
\usepackage{multirow}
\usepackage{ctable}
\definecolor{red1}{hsb}{0, 0.25, 1}
\definecolor{red2}{hsb}{0, 0.5, 1}
\definecolor{red3}{hsb}{0, 0.75, 1}
\definecolor{red4}{hsb}{0, 1, 1}
\newcommand{\comment}[1]{}
\title{Emotion Carrier Recognition from Personal Narratives}
\name{Aniruddha Tammewar, Alessandra Cervone, Giuseppe Riccardi}
\address{
  Signals and Interactive Systems Lab, University of Trento}
\email{\{aniruddha.tammewar, alessandra.cervone, giuseppe.riccardi\}@unitn.it}

\begin{document}

\maketitle
\begin{abstract}
Personal Narratives (PN) - recollections of facts, events, and thoughts from one’s own experience - are often used in everyday conversations.
So far, PNs have mainly been explored for tasks such as valence prediction or emotion classification (e.g. \textit{happy, sad}). However, these tasks might overlook more fine-grained information that could prove to be relevant for understanding PNs.
  In this work, we propose a novel task for Narrative Understanding: Emotion Carrier Recognition (ECR).
  Emotion carriers, the text fragments that carry the emotions of the narrator (e.g. \textit{loss of a grandpa, high school reunion}), provide a fine-grained description of the emotion state. We explore the task of ECR in a corpus of PNs manually annotated with emotion carriers and investigate different machine learning models for the task.
  We propose evaluation strategies for ECR including metrics  that can be  appropriate for different tasks.
  
\end{abstract}

\section{Introduction}
\label{sect:intro}

\begin{table*}
\setlength\tabcolsep{1.5 pt}
\centering
\caption{A small text fragment from a PN annotated with emotion carriers. The first row reports the original German words from the PN, the second row shows the corresponding English translation, while the third row shows the annotations. The annotation is performed by 4 annotators, thus for each token, there are 4 IO labels. For the token ``Familie'' the annotation is I\textbar I\textbar I\textbar O, which means that the first three annotators classified it as \textit{I} while the forth as \textit{O}. The intensity of the red color in the background for the PN fragment also highlights the number of annotators who annotated the token (from lightest for 1 annotator to the darkest for all annotators).}
\vspace{-2mm}
\scalebox{0.85}{
\begin{tabular}{llllllllllll}
\textbf{PN fragment:} & Und & ähm & die & Gefühle & dabei & waren & dass & man & sich &  \\
\textbf{Gloss:} & And & um & the & feelings & there & were & that & you & yourself &  \\
\textbf{Annotation:} & O\textbar O\textbar O\textbar O & O\textbar O\textbar O\textbar O & O\textbar O\textbar O\textbar O & O\textbar O\textbar O\textbar O & O\textbar O\textbar O\textbar O & O\textbar O\textbar O\textbar O & O\textbar O\textbar O\textbar O & O\textbar O\textbar O\textbar O & O\textbar O\textbar O\textbar O &  \\ \hline
\textbf{PN fragment:} & einfach & \colorbox{red1}{\textcolor{white}{\textbf{freut}}} & \colorbox{red1}{\textcolor{white}{\textbf{und}}} & \colorbox{red2}{\textcolor{white}{\textbf{glücklich}}} & \colorbox{red1}{\textcolor{white}{\textbf{ist}}} & dass & man & eine & \colorbox{red3}{\textcolor{white}{\textbf{Familie}}} &  &     \\
\textbf{Gloss:} & easy & pleased & and & happy & is & that & you & a & family &  &    \\
\textbf{Annotation:} & O\textbar O\textbar O\textbar O & I\textbar O\textbar O\textbar O & I\textbar O\textbar O\textbar O & I\textbar O\textbar O\textbar I & O\textbar O\textbar O\textbar I & O\textbar O\textbar O\textbar O & O\textbar O\textbar O\textbar O & O\textbar O\textbar O\textbar O & I\textbar I\textbar I\textbar O &  &      \\ \cline{1-10}
\textbf{Translation:} & \multicolumn{11}{l}{\textit{And uh, the feelings were that you are uh just pleased and happy that you have a family ...}}
\end{tabular}}
\label{tab:example}
\vspace{-5mm}

\end{table*}

A Personal Narrative (PN) is a recollection of events, facts, or thoughts felt or experienced by the narrator. People tell PNs in the form of stories to themselves and to others to place daily experiences in context and make meaning of them \cite{lysaker2001schizophrenia}. 
Rich information provided through PNs can help better understand the emotional state of the narrator, thus PNs are frequently used in psychotherapy \cite{angus2004handbook}. 
Often, in psychotherapy sessions, clients are invited by therapists to tell their stories/PNs \cite{howard1991culture}.
Through PNs, clients provide therapists with a rough idea of their orientation toward life and the events and pressures surrounding the problem at hand \cite{howard1991culture}. Nowadays, well-being applications are widely used, making the collection of PNs easier in the form of a journal from clients in digital form.

Automatic Narrative Understanding (ANU) is a growing field of research that aims at extracting different important and useful information from Narratives for target applications \cite{bamman2019proceedings}. Examples of tasks include reading comprehension \cite{chen2018neural}, summarization \cite{nenkova2011automatic}, and narrative chains extraction \cite{chambers2008unsupervised}.
However, although PNs are widely used in psychotherapy, very few ANU tasks have been proposed to analyze PNs from the perspective of well-being.



A deep emotion analysis of the PNs is possible with ANU.
For example, \cite{tammewar2019modeling} work on predicting valence from the PN and found how different text fragments, including not only sentiment words but also words referring to events or people, proved to be useful to predict the valence score of a narrative for machine learning models.
Following up on this analysis, \cite{tammewar-EtAl:2020:LREC} propose \textit{emotion carriers (EC)} as the concepts from a PN that explain and carry the emotional state of the narrator. ECs thus include not only explicitly emotionally charged words, such as ``happy'', but also mentions of people (friends), places (school, home), objects (guitar), and events (party) that carry an emotional weight within a given context. For example, in the narrative in Table \ref{tab:example} the word ``family'' (orig. ``Familie'') is carrying a positive emotion for the narrator. \cite{tammewar-EtAl:2020:LREC} propose an annotation schema for ECs and perform an annotation experiment of German PNs from the Ulm State-of-Mind in Speech corpus \cite{rathner2018state}.
Based on the analysis of the inter-annotator agreement, they find the task complex and subjective.
Nevertheless, the task has potential benefits in many applications, including well-being.

\begin{figure}
\centering
    \includegraphics[ width=0.47\textwidth]{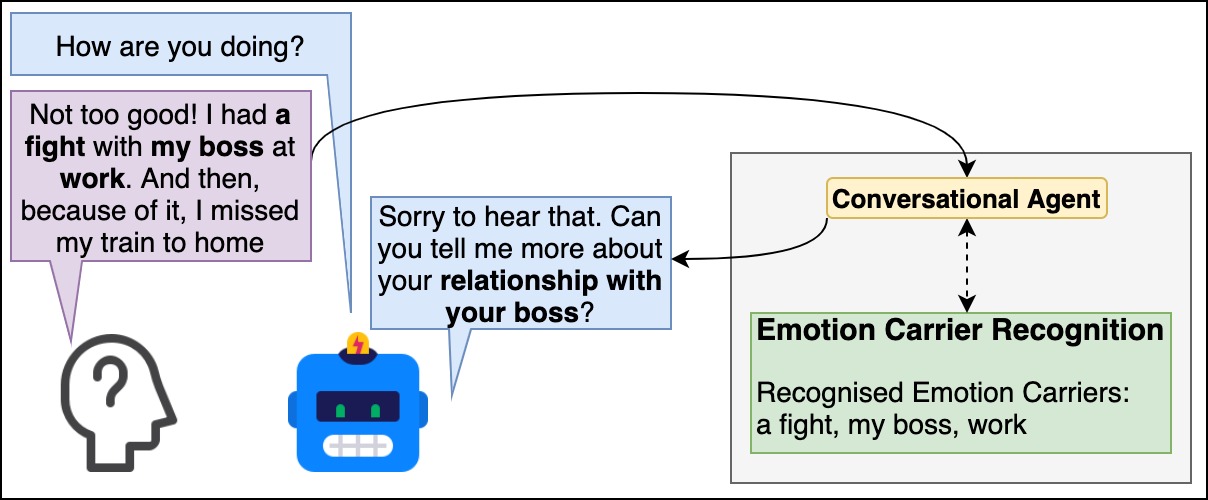}
    \vspace{-4.5mm}
    \caption{A possible application of Emotion Carrier Recognition(ECR): A Conversational Agent (CA), with an ECR  component, first identifies three ECs from the user utterance. One of the ECs "\textbf{my boss}" is then used to generate a response targeted towards eliciting more information from the user, beneficial for a fine-grain description of the user state.
    }
    \label{fig:ECR_application}
    \vspace{-7mm}
\end{figure}

Automatic recognition of ECs could support Conversational Agents (CA) in natural language understanding and generation.
As illustrated in Figure \ref{fig:ECR_application}, a CA could use ECR output to ask follow-up questions based on the recognized ECs. In the example, the user utterance can be considered as a short PN expressing negative sentiment. With current emotion analysis systems, it would be possible to understand that the user is upset about something. This information could be used for example to form a response showing empathy toward the user, e.g. the first part of the response `Sorry to hear that.'.  In addition to emotion categorization, EC recognition (here, the identified ECs include 'a fight', 'my boss', and 'work')  would allow to ask follow-up questions (e.g. `Can you tell me ...') to better understand the relationship between an EC and the narrator's emotional state.

The contributions of this paper are threefold. First, we propose Emotion Carriers Recognition (ECR), a novel task of narrative understanding. Second, to analyze the feasibility of automation of the task, we present different baseline models for ECR relying on bi-LSTM based architectures on an EC annotated corpus \cite{tammewar-EtAl:2020:LREC}. Third, we propose different evaluation strategies for ECR and compare the model's performance to the inter-annotator agreement obtained from humans.

\section{Related Work}
An interesting ANU task, relevant for emotion analysis, is valence (emotional value associated with a stimulus) prediction from spoken PNs \cite{schuller2018interspeech,tammewar2019modeling}.
Another task from emotion analysis that can be borrowed to ANU is Emotion Cause Extraction (ECE), aimed at identifying what might have triggered a particular emotion in a character from a narrative or the narrator himself \cite{lee2010text}.
In ECE, the cause of an emotion is usually a single clause \cite{chen2010emotion} connected by a discourse relation \cite{mann1987rhetorical} to another clause explicitly expressing an emotion \cite{cheng2017emotion}, as in this example from \cite{gui2016event}: ``$<cause>$ \textit{Talking about his honours,} $</cause>$ \textit{Mr. Zhu is so} $<emotion>$ \textit{proud} $</emotion>$.' The focus of ECE, however, has been mainly on text genres such as news and microblogs so far. Applying ECE to the genre of PNs is complicated, given the complex structures of PNs, involving multiple sub-events and their attributes, such as different characters involved, time, and place. In PNs the cause, as well as the emotion, may not be explicitly expressed in a single keyword/clause and multiple keywords might be required to express the same, thus making it difficult to encode it into discourse relation.

Recently, there has been a growing interest in the identification and analysis of \textit{Affective Events (AE)}- activities or states that positively or negatively affect people who experience them, from written texts such as narratives and blogs \cite{ding2016acquiring,ding2018human} (eg. `I broke my arm' is a negative experience whereas `I broke a record' is a positive one.). While AEs are a predefined set of universal subject-verb-object tuples, ECs are text spans from context.

\section{Data}
\label{sect:data}
In this work, we use a corpus of Spoken PNs manually annotated with text spans that best explain the emotional state of the narrator (ECs) following the annotation schema proposed in \cite{tammewar-EtAl:2020:LREC}. The PNs are in German and taken from the Ulm State-of-Mind in Speech (USoMs) corpus \cite{rathner2018state}, which was used and released in the Self-Assessed Affect Sub-challenge, a part of the Interspeech 2018 Computational Paralinguistics Challenge \cite{schuller2018interspeech}.
The USoMs dataset consists of PNs
collected from 100 participants. The participants were asked to recollect two positive and two negative PNs.
The PNs are transcribed manually. Later, PNs from 66 participants, consisting of 239 PNs, were annotated with the ECs by four annotators (25 PNs from the USoMs data were removed because of issues like noise). All the annotators were native German speakers holding Bachelor’s degree in Psychology. They were specifically trained to perform the task. The annotation task involved recognizing and marking the emotion carrying text spans as perceived by the annotators from the PNs.
They were asked to select sequences of adjacent words (one or more) in the text that explain why the narrative is positive or negative for the narrator, focusing specifically on the words playing an important role in the event such as people, locations, objects.

The data can be represented with the \textit{IO} encoding, as shown in the example from Table \ref{tab:example}. We consider the document as a sequence of tokens, where each token is associated with the label \textit{I} if it is a part of an EC, and the label \textit{O} if it is not. A continuous sequence of tokens with label \textit{I} represents an EC.
In the third row \textit{Annotation}, we show the manual annotation by four annotators. It can be observed how the annotators perceive ECs differently, showing the high subjectivity of the task.

In the preprocessing step, we first perform tokenization using the spaCy toolkit \cite{spacy2}. 
Next, we remove punctuation tokens from the data. Based on initial experiments, we found that removing punctuation helps improve the performance of the models.
The number of annotations (ECs) identified by the annotators per narrative varies from 3 to 14 with an average of 4.6. On average, the number of tokens per EC consists of 1.1 tokens for three annotators, while the fourth annotator identified longer segments consisting of 2.3 tokens (avg.). On average, a narrative consists of 704 tokens, while a sentence consists of 22 tokens. The sentence splitting is performed using the punctuation provided in the original transcriptions. 

We find that only 7.3\% of the tokens are assigned the label \textit{I} by at least one annotator. This shows that the classes \textit{I} and \textit{O} are highly imbalanced, which could result in inefficient training of the models. With further analysis, we notice that only 34\% of the total sentences contain at least one EC, while the remaining 66\% sentences do not contain any carrier marked by any annotator.

\section{Task Formulation}
\label{sect:task}
\comment{We propose a novel narrative understanding task of automatic Emotion Carrier Recognition (ECR) from PNs. PNs are collected and used in psychotherapy in different ways.
Diary is a popular life-logging tool for storing memories and daily experiences in the form of PNs, aiding recollection \cite{machajdik2011affective, sellen2007life}. Diaries have shown to improve adherence by increasing the consciousness of the clients about their condition and are effective in gaining deep insights into a client’s well-being. The diary can be used by a therapist to learn about the client’s behavior and routines \cite{gjengedal2010act}. The ECs from the PNs may provide deeper insights about how events, relations, and other elements from the client's life affect his/her mental state.
Our task focuses on the automatic ECR from PNs in the scenario of a mental well-being mobile application with the broader aim of building a conversational agent that can start a conversation with the client based on the PNs shared, to elicit more information about important carriers.}

We pose the recognition of emotion carriers from a given PN as a sequence labeling problem.
The final goal is the binary classification of each token into classes \textit{I} or \textit{O}. As seen in Table \ref{tab:example}, the task of selecting EC text spans is subjective. Each annotator has a different opinion toward the spans to be selected as ECs, so it is challenging to identify an annotation as valid or invalid. For each token, we have annotations from four annotators with \textit{IO} labels. Some annotators may agree on the annotation, but it is infrequent that all four annotators annotate the token as \textit{I}. In the example from Table \ref{tab:example}, the token ``glücklich'' is annotated \textit{I} by two annotators, three annotators agree that the token ``Familie'' is an EC, while a few tokens are marked as EC by only one annotator. In this scenario, it is difficult to provide a hard \textit{I} or \textit{O} label. To tackle this problem, we model the problem of Emotion Carrier Recognition (ECR) as providing scores to the label \textit{I}, representing the likelihood that that token is a part of an EC. Label distribution learning (LDL) \cite{geng2016label} can effectively capture the label ambiguity and inter-subjectivity within the annotators.
We use LDL with a sequence labeling network and the KL-Divergence loss function. The advantage of LDL is that it allows to modeling the relative importance of each label. For evaluation, we use different strategies to select the final \textit{IO} labels.

\section{Model}
\label{sect:Model}
We use sequence labeling architecture relying on biLSTM with attention, 
similar to \cite{shirani2019learning}.
\begin{figure}
    \centering
    \includegraphics[width=0.36\textwidth]{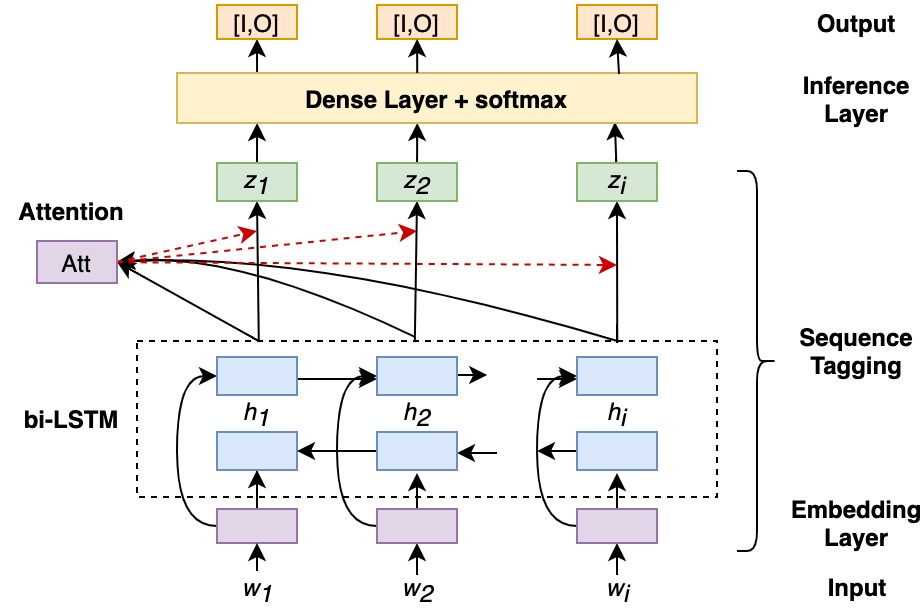}
    \vspace{-1mm}
    \caption{bi-LSTM based DNN architecture for ECR. In the output, [I,O] represent the probabilities for the classes \textit{I} and \textit{O}.}
    \label{fig:Architecture}
\vspace{-0.6cm}
\end{figure}
As shown in Figure \ref{fig:Architecture}, the input text is first passed through the embedding layer to obtain the word embedding representation for each token. We use 100-dimensional pre-trained GloVe \cite{pennington2014glove} embeddings. To encode the sequence information, we then use two stacked bidirectional LSTM layers with a hidden size of 512. We also use attention mechanism \cite{zhang2017position} along with the bi-LSTM, where attention weights $a_i$ represent the relative contribution of a specific token to the text representation. We compute $a_i$ at each output time $i$ as follows:
\vspace{-2mm}
\begin{equation}
\vspace{-2mm}
    a_i = softmax(v^T tanh(W_hh_i + b_h))
\vspace{-2mm}
\end{equation}
\vspace{-2mm}
\begin{equation}
\vspace{-1mm}
    z_i = a_i.h_i
\end{equation}
where $h_i$ is encoder hidden state and $v$ and $W_h$ are learnable parameters of the network. The output $z_i$ is the element-wise dot product of $a_i$ and $h_i$.

Finally, the output is passed through the inference layer consisting of two fully connected layers with 50 units each, and a softmax layer to assign probabilities to the labels for each word. We also use layer normalization and two dropout layers with a rate of 0.5 in the sequence and inference layers \cite{ba2016layer}

During training, we use the Kullback-Leibler Divergence (KL-DIV) as the loss function \cite{kullback1951information}
and the Adam optimizer \cite{kingma2014adam} with the learning rate of 0.001.

\section{Experiments}
\label{sect:experiments}
As described in Section \ref{sect:data}, there is class imbalance in the data. The class \textit{I} tokens are very infrequent compared to the class \textit{O}.  
This may result in a bias toward class \textit{O} in the classifier. Another problem we have to deal with is the length of the narratives. The narratives are very long with an average length of 704 tokens. Standard machine learning and bi-LSTM based architectures are not efficient in dealing with very long contexts.

To address these challenges, we experiment with different levels of segmentation of the narratives and apply strategies to select proper train and test sets. We train and test the sequence-labeling models at narrative and sentence levels. In the \textbf{narrative} level, we consider the entire narrative as one sequence, while in the \textbf{sentence} level, we consider one sentence as a sequence. In this way, we analyze how the length of a sequence affects the performance of the model. Also, note that at the sentence level, the model does not have access to other parts of the narrative. We study how limited access to context affects performance.

The sentence-level sequences are further considered in two ways : 1) \textbf{SentAll:} all the sentences are considered 2) \textbf{SentCarr:} only sentences containing at least one EC are considered. \textit{SentCarr} reduces the class imbalance as we remove all sentences that do not contain any token tagged as class \textit{I}. In a real-world scenario, we would have to extract carriers from the entire narrative or all the sentences, as we do not know beforehand which sentences contain the carriers. Thus, we use \textit{SentCarr} only for training, but in the test set \textit{SentAll} is used.

We also experiment with another sequence labeling model based on \textbf{Conditional Random Fields (CRF)} \cite{lafferty2001conditional}, a widely used machine learning algorithm for sequence-labeling problems in NLP, such as Part of Speech tagging.
For the CRF model, we use the context window of $\pm 3$ with features such as the token, its suffixes, POS tag, prefix of POS tag, sentiment polarity.

\section{Evaluation}

\begin{figure*}[htp]
  \subfloat[Model's Output]{
	\begin{minipage}[c]{0.48\textwidth}
	   \centering
	   \includegraphics[width=1\textwidth]{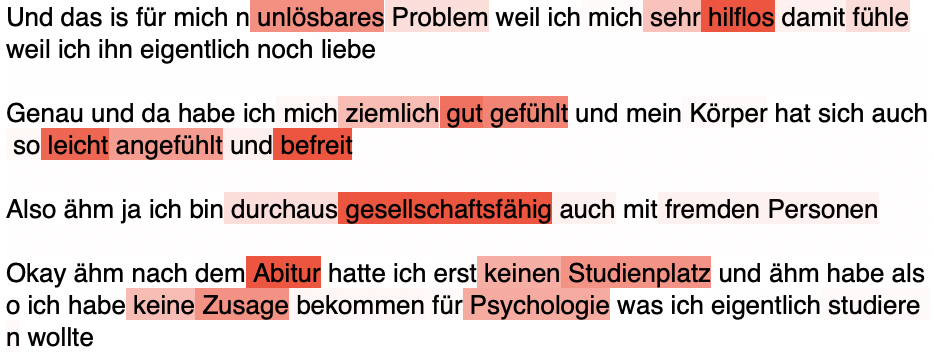}
	   \vspace{-0.5cm}
	\end{minipage}}
 \hfill 	
  \subfloat[Ground Truth]{
	\begin{minipage}[c]{0.48\textwidth}
	   \centering
	   \includegraphics[width=1\textwidth]{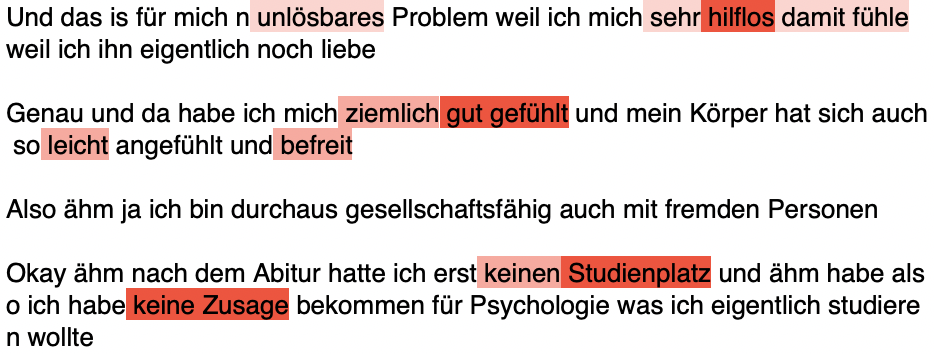}
	   \vspace{-0.50cm}
	\end{minipage}}
\vspace{-4mm}
\caption{Heatmap of sentences from narratives annotated with emotion carriers; highlighting tokens with model’s output and ground truth probabilities. Notice the wider range of scores in the Model's Output as compared to only four possible scores in the Ground Truth.}
\vspace{-4.5mm}
\label{Fig:heatmap}
\end{figure*}

In this section, we propose different evaluation strategies for the ECR task. Note that even though our model is trained to predict the probability distribution of the classes, our final goal is to assign one of the two classes from \textit{I} and \textit{O}. For all evaluations, for the ground truth, we consider that a token is annotated (i.e. \textit{I} in the \textit{IO} tags) if at least one of the annotators has annotated it. Similarly, for the output, we consider the output as \textit{I} if the probability assigned crosses the minimum threshold of 0.25, which is equivalent to one of four annotators tagging the token as \textit{I}.
In all evaluations, we do five-fold cross-validation with the leave one group (of narrators) out (LOGO) strategy. For each training session, we split the data into train, dev, and test sets without any overlap of narrators in the three sets.


\subsection{Token Level}
The token level evaluation measures the performance of predicting \textit{I} or \textit{O} class for each token in a sequence. We use this metric to evaluate our models with different data segmentations. We are concerned more about the prediction of the class \textit{I}, as we are interested in applications of ECR such as Figure \ref{fig:ECR_application}, where it is important to find one or more important carriers to start a conversation with the narrator. Thus, we show the F1 score of class \textit{I} and weighted average (micro) of F1 score of \textit{I} and \textit{O}.

As discussed earlier, considering a real-world scenario, we need the model to perform well on the \textit{SentAll} or \textit{Narrative} data. In Table \ref{tab:token-results}, we find that the model trained on \textit{SentCarr} performs best on the \textit{SentAll}. For further evaluation we use this model, thus recognition would be done on the sentences and not the entire narrative at once. Note that the performance of the \textit{Narrative} strategy is only slightly worse, suggesting that the task is not affected much by the length of the context available. The CRF model is the worst performing one.

Using the \textit{SentCarr} model, we extract the continuous sequences of tokens that are tagged as \textit{I}. These text spans are considered as the ECs recognized by the model. In the metrics in the next section, we evaluate the model by comparing this set of carriers with the set of manually annotated reference carriers. 
\begin{table}
\setlength\tabcolsep{3 pt}
\centering
\centering
\caption{Results of bi-LSTM based models with different data segmentation.
Notice how the the number of data-points vary as we change the segmentation.
}
\vspace{-1mm}
\scalebox{1.0}{
\begin{tabular}{|l|l|l|l|l|} 
\hline
\multicolumn{3}{|l|}{\textbf{Data} } & \multicolumn{2}{l|}{\textbf{Results (F1)(std)}} \\ 
\hline
 \textbf{Segmentation}  & \textbf{\#train}  & \textbf{\#test}  & \textbf{class-I}  & \textbf{micro}  \\ 
\hline
SentCarr & 1737 & 367 & 53.2(4.7) & 93.7(0.8) \\ 
\hline
\begin{tabular}[c]{@{}l@{}}\textbf{train: SentCarr}\\ \textbf{test: SentAll} \end{tabular} & 1737 & 1533 & \textbf{34.9(3.4)} & \textbf{96.6(0.5)} \\ 
\hline
SentAll & 6378 & 1533 & 31.2(3.9) & 96.6(0.5) \\ 
\hline
Narrative & 191 & 48 & 34.6(5.0) & 96.7(0.4) \\ 
\hline
\begin{tabular}[c]{@{}l@{}}CRF(SentCarr;\\SentAll)\end{tabular} & 1674 & 1582 & 34.2(4.1) & 96.2(0.3) \\
\hline
\end{tabular}}
\label{tab:token-results}
\vspace{-4mm}
\end{table}



\subsection{Agreement Metrics}
\label{sect:agreement-metrics}
\begin{table}
\setlength\tabcolsep{2 pt}
\centering
\caption{Evaluation based on the agreement metrics (positive agreement) with different parameter configurations. For each configuration, the corresponding inter-annotator agreement (IAA) score is in the last column (in terms of F1 score).[\textbf{Parameters}:(Matching strategy:Exact, Partial); (Position: considered(T), agnostic(F)); (lexical level: token, lemma)]}
\vspace{-0.5mm}
\hspace{-0.3cm}
\scalebox{1.0}{
\begin{tabular}{|l|l|l|l|l|l|} 
\hline
\textbf{sr}  & 
\textbf{Parameters}  & \textbf{Prec(std)}  & \textbf{Recall(std)}  & \textbf{F1(std)}  & \begin{tabular}[c]{@{}l@{}}\textbf{IAA}\\\textbf{(F1)} \end{tabular}  \\ 
\hline
a & \begin{tabular}[c]{@{}l@{}}Exact, F, token\\ (w/ stopwords) \end{tabular} & 32.6(3.3) & 52.1(3.6) & 40.0(2.9) & 25.2 \\ 
\hline
b & \begin{tabular}[c]{@{}l@{}}Exact, F, token \end{tabular} & 42.3(4.6) & 67.2(4.2) & \begin{tabular}[c]{@{}l@{}}51.7(4.0)\\\end{tabular} & NA \\ 
\hline
c & \begin{tabular}[c]{@{}l@{}}Partial, T, token \end{tabular} & 37.4(4.0) & 51.3(0.4) & 43.1(2.7) & 32.0 \\ 
\hline
d & \begin{tabular}[c]{@{}l@{}}Partial, F, token \end{tabular} & 59.4(5.5) & 83.6(4.7) & 69.2(3.5) & 39.9 \\ 
\hline
e & \begin{tabular}[c]{@{}l@{}}Partial, F, lemma \end{tabular} & 61.8(6.4) & 86.5(4.2) & 71.8(4.3) & 40.3 \\
\hline
\end{tabular}}

\vspace{-5mm}
\label{tab:agreement_metrics}
\end{table}
We evaluate the performance of the models using the metrics that were used to evaluate the inter-annotator agreement between the four annotators (pair-wise) in \cite{tammewar-EtAl:2020:LREC},
based on the \textit{positive (specific) agreement} \cite{fleiss1975measuring}. This evaluation is important as it compares the performance of the system with the inter-annotator agreement, which can loosely be considered as human performance.

We also explore the different criteria to decide whether two spans match or not, as used in the original metrics. The evaluations are based on \textit{Exact Match}, where the two carriers match if they are exactly the same and \textit{Partial/soft Match}, where the token overlap of the two carriers wrt the reference carrier is considered. Other parameters in the matching criteria include \textit{position of the carrier in the narrative}, which has two possibilities for matching of the two candidates. \textit{Position agnostic}, where position does not matter for matching and \textit{position considered}, where the two spans have to be at the same position. Another criterion is based on the matching of \textit{tokens} vs \textit{lemmas}.
We remove stopwords from the annotation as we are interested in the content words.

\textbf{Results:} Table \ref{tab:agreement_metrics} summarizes the evaluation using the Agreement metrics. As expected, with the loosening of the matching criteria, the results improve. A similar trend is observed in the inter-annotator agreement. When we move from \textit{a} to \textit{b}, we are removing the stopwords from the predicted and reference carriers. This improves the results significantly. The reason behind this is the fact that the reference annotations, which were also used for training the model, as mentioned earlier, contain all the tokens that are tagged by at least one annotator. As noticed by \cite{tammewar-EtAl:2020:LREC}, in the annotations, one of the annotators usually annotates longer spans than others. They also observed that many annotations also contain punctuation and stopwords. To understand this issue, let us consider an example of concept annotation. For a concept like a printer, the annotators could select spans ‘with the printer’, ‘the printer’ or just ‘printer’. With our strategy for creating reference annotations, we end up selecting the longest span "with the printer" which contains stopwords like \textit{with, the}. However, this might not be the case in the model's output (as the training data also contain concepts marked by only one annotator). To reduce this effect, one way is to remove the stopwords  (strategy b) and another is to use the partial match (strategy c). While both strategies improved the scores, the improvement with strategy b is more significant than with strategy c. We notice a significantly large jump in the model's performance from \textit{c} to \textit{d}, compared to the inter-annotator agreement. Our inituition is that this could be because the model is trained at the sentence level, thus the position in the narrative is not taken into consideration, resulting in recognition of multiple occurrences of the same carrier. Additionally, the performance further improves when we match lemmas instead of tokens (from \textit{d} to \textit{e}).

\textbf{Analysis:} Figure \ref{Fig:heatmap} shows example sentences from a test set with a heatmap showing the model’s predicted score and ground truth probabilities for each token. 
In most cases, the probability distribution in the model's output seems to follow similar trends to that of the ground truth probabilities. We also observe frequent cases of false positives, where the model assigns a high probability to class \textit{I} even when the ground truth label is \textit{O}, as can be seen in the third example. This behavior could be a result of training the model at the sentence level with the \textit{SentCarr} strategy, where all the sentences in the training set contain at least one EC, biasing the model towards that distribution. 

To study if the model is biased towards the recognition of ECs with sentiment words (angry, joy) versus content words (internship, parents) in ECs, we find the sentiment and content carriers using the polarity score assigned by the textblob-de\footnote{\url{https://tinyurl.com/textblob-de}} library. We find the mean of fraction of content ECs per narrative to be 60\% in the reference while 64\% in the prediction, suggesting a minimal bias in the model.


\section{Conclusions}

We proposed Emotion Carriers Recognition (ECR), a novel task to recognize the text spans that best explain the emotional state of the narrator from personal narratives. The proposed task allows to have a fine-grained representation of the emotional state of the narrator by recognizing relevant text fragments, including mentions of events, people, or locations. We presented different baseline models to address the task, and evaluated them using both token-level and agreement metrics. We compared our best model performance with the inter-annotator agreement and found that our model agrees well with the annotators.
We believe this task could be useful as a first step towards a richer understanding of emotional states articulated in personal narratives. As future work, we plan to investigate the integration of a ECR model into a conversational agent, supporting well-being.

\section{Acknowledgements}
The research leading to these results has received funding
from the European Union – H2020 Programme under grant
agreement 826266: COADAPT.

\balance
\bibliographystyle{IEEEtran}
\bibliography{mybib}


\end{document}